\documentclass[11pt,a4paper]{article}
\usepackage[hyperref]{emnlp2020}
\usepackage{times}
\usepackage{latexsym}
\usepackage{times}
\usepackage{soul}
\usepackage{amsmath, amssymb}
\usepackage{bm}
\usepackage{latexsym}

\usepackage{booktabs}
\usepackage{tabularx}
\usepackage{enumitem}
\usepackage{graphicx,subfigure}

\newcommand{\contribsize}{\ensuremath{\gamma}}

\newif\ifcomments
\ifcomments
    \providecommand{\jens}[2][]{{\protect\color{blue}{[Jens:\textbf{#1} #2]}}}
    \providecommand{\changed}[1]{{\protect\color{red}{#1}}}
\else
    \providecommand{\jens}[2][]{}
    \providecommand{\changed}[1]{#1}
\fi

\usepackage{microtype}

\aclfinalcopy 

\title{Differentially Private Language Models Benefit from Public Pre-training}

\author{
  Gavin Kerrigan$^*$ \\
  University of California, Irvine \\
  \texttt{gavin.k@uci.edu} \\\AND
  Dylan Slack$^*$ \\
  University of California, Irvine \\
  \texttt{dslack@uci.edu} \\\And
  Jens Tuyls\thanks{*All authors contributed equally.} \\ 
  Princeton University \\
  \texttt{jtuyls@princeton.edu} 
  }

\date{}

\begin{document}
\maketitle

\begin{abstract}
   Language modeling is a keystone task in natural language processing. When training a language model on sensitive information, differential privacy (DP) allows us to quantify the degree to which our private data is protected. However, training algorithms which enforce differential privacy often lead to degradation in model quality. We study the feasibility of learning a language model which is simultaneously high-quality and privacy preserving by tuning a public base model on a private corpus. We find that DP fine-tuning boosts the performance of language models in the private domain, making the training of such models possible.\footnote{\texttt{https://github.com/dylan-slack/ Finetuning-DP-Language-Models}}
\end{abstract}

\section{Introduction}

Language modeling, the task of assigning a probability to sequences of words, is a key problem in natural language processing. 
Modern language models are data-driven, relying on a large corpus of text. Many such models are trained on corpora from a specific domain, such as Wikipedia or news articles~\cite{radford2019language}. These models often suffer from generalization issues when used to model language from a different domain. This motivates the use of model fine-tuning, in which the weights of a pre-trained language model are tuned by gradient descent on a second dataset of interest~\cite{radford2019language, Devlin2019BERTPO, Liu2019RoBERTaAR}. 

 In some cases, we would like to fine-tune our model with respect to a dataset containing private information. As such, there is an obligation to preserve the privacy of individuals who contribute text to the private training corpus. For example, training a medical chat-bot may require learning a language model from transcribed patient-doctor conversations; it would be critical that this model not expose sensitive information about the patients whose conversations are used as training data.  In recent years, differential privacy (DP) has been a key quantitative measure of privacy which allows one to use aggregate statistical information about a dataset while preserving the privacy of its individual datapoints. 
 
 In the case of language modeling, we are interested in preserving the privacy of individuals who contribute text to a private corpus. As each individual who contributes to this dataset could potentially contribute several sentences, our notion of privacy is \textit{group differential privacy} \cite{dwork_privacybook}, in which all sentences from a single individual are grouped. In practice, group DP is equivalent to DP with re-scaled parameters. A potential limitation of this approach is that the number of contributed sentences may not be uniform over users, leading to sub-optimal bounds on the privacy guarantee. There has been some success in directly training differentially private language models, but these often require access to large datasets in order to achieve a reasonable level of quality \cite{recurrent_DP}. Other work has trained a differentially private base model which was then fine-tuned through active learning on a non-private dataset  \cite{zhao2019improving}.
 
 We instead train a non-private base model on a large, public dataset, which we proceed to fine-tune on a private out-of-distribution dataset through differentially private stochastic gradient descent (DPSGD) \cite{2016arXiv160700133A}. By doing so, we successfully train a high-quality model which is differentially private with respect to our tuning dataset. Our experimental results show that DP fine-tuning not only boosts the performance of DP language modeling, but makes it possible. 

\section{Related Work}

Training a feedforward neural network with DP is achievable through the popular DP-SGD algorithm \cite{2016arXiv160700133A}. However, this method may lead to significant decreases in the accuracy (or other metrics) of the resulting model. Recent work considers the use of metric privacy for language modeling \citep{fernandes2019generalised,feyisetan2020privacy}, which is a relaxation of differential privacy where noise is instead added to the vector embedding of a word. We leave the exploration of metric privacy for the private fine-tuning task as a direction for future work.

Many high-quality language models rely on some form of recurrent neural architecture, such as RNNs or LSTMs \cite{DBLP:journals/corr/abs-1808-03314, 10.1162/neco.1997.9.8.1735}. In \cite{recurrent_DP}, the authors develop a method for training such models while achieving differential privacy. However, this approach requires a large private dataset, and the mechanisms to achieve privacy lead to a significant decrease in model quality.

In \cite{zhao2019improving}, the authors attempt to train a language model which is simultaneously differentially private and of high quality. The first solution proposed in \cite{zhao2019improving} is to fine-tune the language model with publicly available data, but as this public data is likely distributed differently than the private data, the resulting model is likely mistuned. The second proposed approach is to augment the training data by actively selecting non-private data instances. This effectively reduces the privacy cost incurred during each training step, but still requires training with potentially out-of-distribution data. 

In contrast, our work begins with a pre-trained model which only has access to publicly available data. This base model is then fine-tuned through DPSGD on our private domain of interest, resulting in a model that is both differentially private and tuned with respect to our protected dataset. By tuning a pre-trained public model, we achieve higher quality models without incurring any additional costs to our privacy budget.

\section{Approach}

Let $\mathcal{D}$ be a publicly available corpus, and $\mathcal{P}$ be a protected corpus whose contents we would like to protect the privacy of. Denote by $\mathcal{X}$ the fixed, shared vocabulary of these corpora. 
At a high level, our approach is to first train a language model $M_\mathcal{D}: \mathcal{X}^n \to [0, 1]$. In practice, we choose a feedforward architecture for $M_\mathcal{D}$ due to limited computing resources. We fine-tune this model with respect to $\mathcal{P}$ by using the DPSGD algorithm \cite{2016arXiv160700133A} on batches of sentences from $\mathcal{P}$.

\subsection{$(\epsilon, \delta)$ Differential Privacy}

Intuitively, an algorithm is $(\epsilon, \delta)$-DP if the output of the algorithm cannot be used to probabilistically determine the presence of a single instance in the database by more than a factor of $\exp(\epsilon)$. We additionally allow this constraint to be violated with probability $\delta$, with $\delta$ typically being small\footnote{Some authors recommend a value of $10^{-5}$ \cite{2016arXiv160700133A}.}.

In the case of language modeling, an individual $i$ may possibly contribute $s_i \geq 1$ sentences to the private training corpus. To maintain the privacy of said individual, we require that our algorithm satisfy $s_i$-\textit{group differential privacy}, meaning our algorithm cannot be used to determine the presence or absence of $s_i$ sentences in the dataset. However, $(\epsilon, \delta)$ $s_i$-group DP is equivalent to $(\epsilon/s_i, \delta)$-DP \cite{dwork_privacybook}. Hence, it is sufficient to consider the somewhat unintuitive notion of preserving the privacy of individual sentences in the training set. Any mechanism satisfying $(\epsilon, \delta)$-DP on individual sentences will then satisfy $(\epsilon/\contribsize{}, \delta)$-DP with respect to contributing individuals, where $\contribsize{} = \max_i \{ s_i \}$.
Formally, an algorithm $\mathcal{A}$ satisfies $(\epsilon, \delta)$-DP if for all datasets $\mathcal{D}_1, \mathcal{D}_2$ differing by at most one instance, and for any set $S$, we have

\[\mathbb{P} \{\mathcal{A} (\mathcal{D}_1) \in S \} \leq \exp(\epsilon) \mathbb{P} \{\mathcal{A} (\mathcal{D}_2) \in S  \} + \delta \]

Smaller $\epsilon$ values indicate a stronger privacy guarantee. We typically think of $S$ being some query on the outcome of $\mathcal{A}$. A more complete treatment of differential privacy is available in \cite{dwork_privacybook}.

\subsection{Differentially Private Fine-tuning}

Differential privacy is achieved in SGD by adding appropriately scaled noise to the gradient of the loss function.
In particular, we fix a noise scale $\sigma^2 \in \mathbb{R}$ and a gradient clipping level $C\in \mathbb{R}$. For a batch of size $L$, our loss function is given by $\mathcal{L}(\theta) = \frac{1}{L} \sum_i \mathcal{L}(x_i ;  \theta)$. For each $x_i$ in our batch, we compute the clipped gradient $g(x_i)$ by scaling the gradient of the loss at $x_i$ to have $\ell_2$ norm at most $C$ \cite{dwork_privacybook}.

\[g(x_i) = \frac{1}{\max \{1, ||\nabla_\theta \mathcal{L}(x_i; \theta)||_2 / C \}} \nabla_\theta \mathcal{L}(x_i; \theta) \]

We add appropriately scaled zero-mean Gaussian noise to our gradients:

\[\widetilde{g}(x_i) = g(x_i) + \mathcal{N}(0, \sigma^2 C^2 I) \]

Our gradient signal used in training is then the average of $\widetilde{g}(x_i)$ over a given mini-batch, which we use to determine a descent direction as in SGD. Note that our noisy gradient is equal to the true gradient in expectation, as we add mean-zero noise. 

As our access to the private data is done entirely in the calculation of $g(x_i)$, with appropriately chosen parameters this method guarantees our algorithm respects our specified level of privacy.


For given noise $\sigma$, we can determine an acceptable privacy violation level $\delta \ll 1$ and compute the resulting privacy parameter $\epsilon$ through the composition theorem proved in \cite{2016arXiv160700133A}. In appendix \ref{DP-guarantees} \ref{fig:eps_delta_refined}, we plot the $(\epsilon, \delta)$-privacy guarantees for various settings of $\sigma$. As expected, for a fixed $\delta$, more noise (greater $\sigma$) results in a tighter privacy guarantee (smaller $\epsilon$).

Throughout this section, we have assumed a maximum individual contribution size of $\contribsize{} = 1$. When $\contribsize{} > 1$, the only necessary change is a post-processing scaling of  $\epsilon \mapsto \epsilon / \contribsize{}$, as $\epsilon$ is computed based on parameters which are independent of $\contribsize{}$. 

\begin{figure*}
\subfigure[Small Language Model]{\includegraphics[width =.5\textwidth]{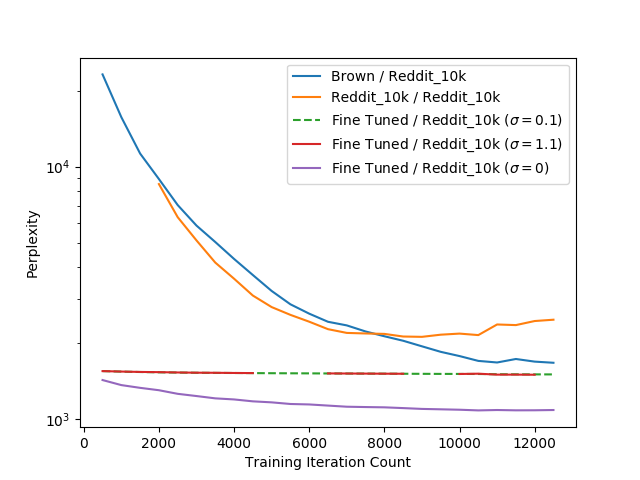} \label{fig:test_pp}}
\quad
\subfigure[Large Language Model]{\includegraphics[width =.5\textwidth]{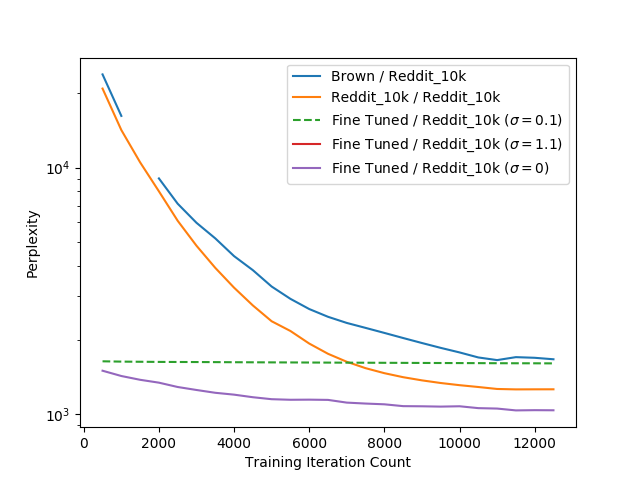}
\label{fig:test_pp_big}}

\caption{\textbf{Test-set perplexity as a function of training iterations for the small (a) and large (b) language models.} The legend indicates train-set / evaluation set, with $\sigma$ being the noise scale used in differentially private training. The fine-tuned models are trained on the Brown corpus and tuned on the Reddit dataset. The graph for $\sigma = 1.1$ for the large language model is not visible since all perplexity values are infinity. Note: the graphs are truncated to the first epoch of training.  Perplexities change marginally after this point.}
\label{fig:exps_small_large}
\end{figure*}

\section{Experimental Results}

\subsection{Datasets}

For our public dataset, we choose the Brown corpus \cite{francis79browncorpus}, as it is a fairly large corpus designed to represent modern English. For our private dataset, we used the Reddit comments dataset \cite{reddit_data}. While this corpus is not truly private, we felt it represented the type of language data one might be interested in protecting -- written language generated by individual users which likely contains personal information. We randomly select a subset of $10$k comments for private training data and $5$k comments for development and testing. \changed{For more details, see Appendix~\ref{appendix:dataset}.}

 \subsection{Models and Evaluation}
 \label{sec:evaluation}

For our language models, we consider two feed-forward architectures: a small network and a large network, each with three hidden layers, but with varying numbers of nodes (see appendix \ref{architectures} for details). For both architectures, we train three baseline models:

\begin{itemize}[nosep]
    \item A non-private model trained only on the public corpus.
    \item A non-private model trained only on the private corpus.
    \item A non-private model pre-trained on the public corpus, and fine-tuned on the private corpus.
\end{itemize}

For each architecture, we compare these baseline models to a private model which is pre-trained on the public corpus and fine-tuned on the private corpus. For the private models, we hold $\delta=1e-5$ and set gradient clipping to $1.0$.  We train each private model with $\sigma=1.1$ and $\sigma=0.1$. \changed{Also, we fine-tune OpenAI's pre-trained GPT-2~\cite{Radford2019LanguageMA} non-privately on both Brown and Reddit.} For each model, we report the perplexity scores.

\subsection{Results}

\paragraph{GPT-2 Fine-tuning }

The GPT-2 model fine-tuned for three epochs on the Brown training data set scored $40.0$ perplexity on the held out test set.  The GPT-2 model fine-tuned for the same time on the Reddit training data set scored $45.14$ on the held out test set.

\paragraph{Small Feedforward Neural Network}

Next, we trained and evaluated a smaller feedforward neural network on the evaluation schema from section \ref{sec:evaluation}.
Figure \ref{fig:test_pp} shows the test-set perplexity for each of our models as a function of training iterations. We observe that each of the base non-private models converges at roughly the same rate, but the models trained on the Brown corpus converge to a lower perplexity than those trained on the Reddit corpus. We also note that the fine-tuned models achieve a significantly lower perplexity in fewer iterations, even with the inclusion of differential privacy mechanisms. The increase in perplexity seen in the base Reddit model may be indicative of overfitting. 

\paragraph{Large Feedforward Neural Network}

Next, we train and evaluate a large feedforward neural network model.  The results can be found in figure \ref{fig:test_pp_big}. We found that the larger models performed mostly similar to the smaller ones. However, the larger model does significantly outperform its smaller counterpart when trained and evaluated on 10,000 comments sampled from the Reddit dataset. This can be seen when comparing figure \ref{fig:test_pp} and \ref{fig:test_pp_big}. The ``Reddit\_10k / Reddit\_10k" curve reaches a much lower value much sooner for the larger model. Another difference is that the larger model was not able to get finite perplexity values when fine-tuned on Reddit\_10k in a differentially private way with noise set to 1.1, while the smaller model was able to do this. 

\section{Analysis}

\paragraph{Finetuning improves DP perplexity} We summarize the perplexities of our final small and large models in table~\ref{tab:finaltessetperplexities} in the appendix. A $\sigma^2$ of zero indicates non-private training while a $\sigma^2 > 0$ indicates private training where privacy increases with larger $\sigma^2$.  We additionally provide the $\epsilon$ values for the private models in figure \ref{fig:epsilon_vs_perplexity}.  The perplexity scores for both the small and large feedforward language models are orders of magnitude worse than the GPT-2 models indicating that they are not competitive with state of the art language models. 

However, our results indicate that pre-training may significantly improve the perplexity of a differentially private language model.  We were unsuccessful in training a differentially private model on the Reddit data alone, as all models tested gave unreasonably high perplexities (i.e. useless models). When DP fine-tuning was used to create a private language model for this domain, our small model outperformed the baseline models \changed{(except for its non-private equivalent)}. This indicates that pre-training may be highly valuable in facilitating the training of DP language models. 

\paragraph{Qualitative Analysis}

We provide a sample of sentences generated from models fine-tuned on the Reddit $10k$ data set in table \ref{tab:generatedsentences} in the appendix.  

Aside from the state of the art GPT-2 model, both the small and large feedforward neural networks are not able to generate sentences that are coherent. Additionally, there is not a discernible difference between the various levels of private fine-tuning.  This is likely because feedforward neural networks are not strong language models. \changed{We do see the pre-training benefits for privacy with such models.} 

\section{Conclusions}
Training neural models with differential privacy often significantly degrades model performance. However, differential privacy could prove crucial when doing language modeling on private datasets. Our work shows that DP fine-tuning not only boosts the performance of DP language modeling, but makes it possible. 
We also compared our experiments across two different model sizes and found that increasing the model size while decreasing the number of training epochs does not significantly impact the results in the differentially private transfer learning scenario. 
Future research could experiment with stronger model architectures (e.g., LSTM's, transformers) instead of regular feedforward neural networks, as well as train models longer in order to increase performance.


\bibliography{acl20xx}
\bibliographystyle{acl_natbib}

\newpage

\appendix
\section{Architectures}
\label{architectures}

We consider two language model architectures.  We first use a feedforward neural network as our language model with three hidden layers consisting of $500$, $250$, and $50$ nodes respectively (``small'' language model).  Recent work suggests large language models may produce better results more quickly than smaller models \cite{Li2020Efficient}.  Though the mentioned work considers transformer models, we also investigate training a larger feedforward neural network with three hidden layers consisting of $10,000$, $5,000$, and $1,000$ nodes (``large'' language model) in hopes to speed up differentially private training and gain better performance. 

For both models, we consider $20$ previous tokens.  We trained the public models using the Adam optimizer with a learning rate of $1e-3$. To train the private models we used the DPSGD optimizer from~\cite{pyvacy}. We used the ReLU activation function on all nodes and the softmax function on the output layer.

Lastly, we trained the small language model for $5$ epochs during pre-training and $5$ epochs during fine-tuning.  We trained the large language model for $2$ epochs during pre-training and $2$ epochs during fine-tuning

\section{$(\epsilon-\delta)$-Privacy Guarantees} \label{DP-guarantees}

\begin{figure}[h!]
    \centering
    \includegraphics[scale=0.5]{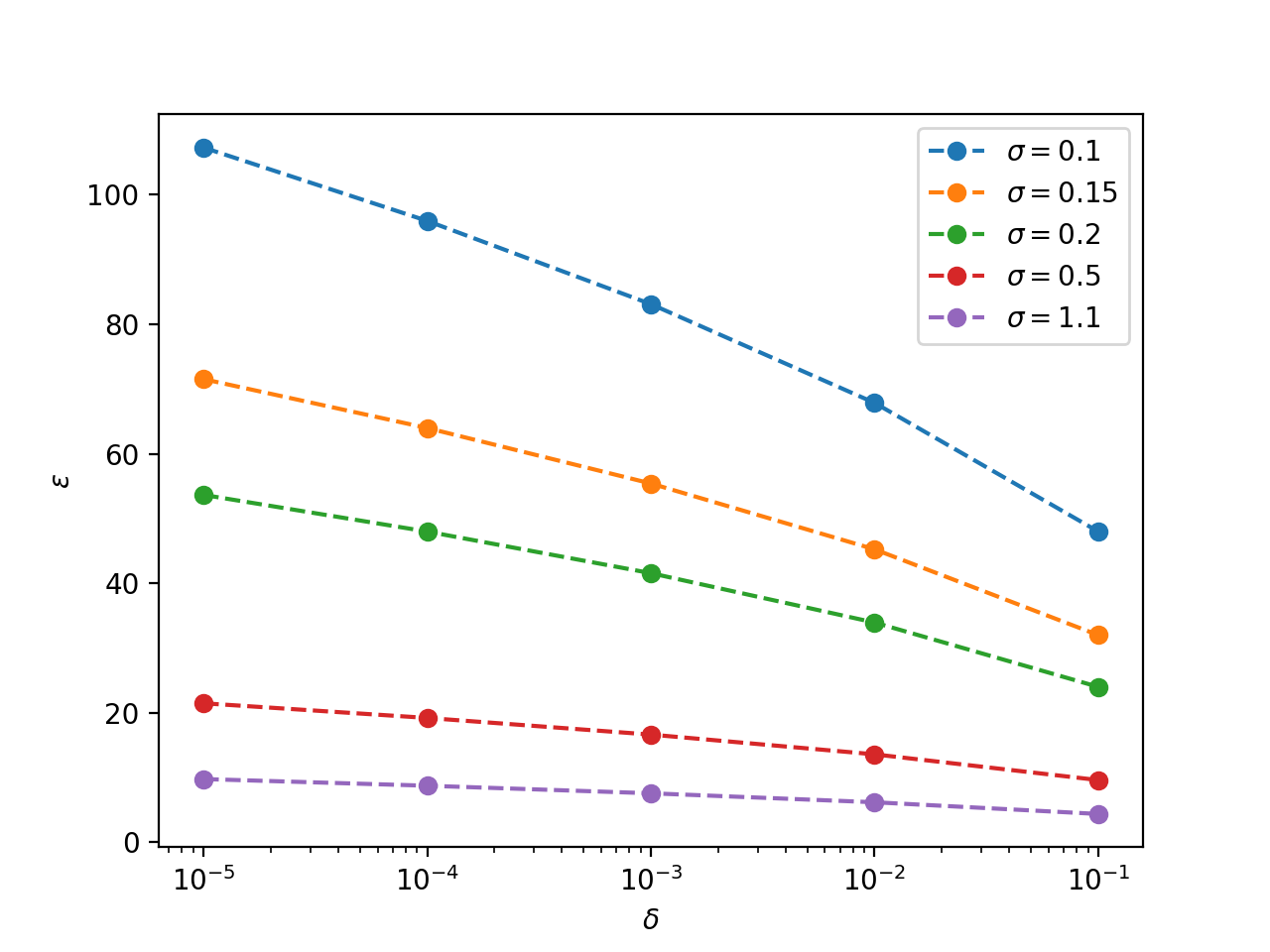}
    \caption{$(\epsilon, \delta)$-privacy guarantees for $q = 10^{-3}, T = 10^5$, computed using the moments accountant \cite{2016arXiv160700133A}. Here, $\sigma$ is a noise-scale parameter specified by the user. This helps us to select a noise scale appropriate to a given application setting. }
    \label{fig:eps_delta_refined}
\end{figure}

\section{Dataset Sizes}\label{appendix:dataset}

In figure \ref{fig:datasetsize}, we provide the number of tokens used for training in  each data set.

\begin{figure}[h!]
\centering
\begin{tabular}{lll}
\textbf{Dataset}    & \textbf{Tokens (train)} & \textbf{Tokens (test)} \\ \hline
Reddit\_10k & \changed{689,763} & \changed{344,120}   \\ 
Brown      & \changed{693,683} & - \\ 
\end{tabular}
\caption{The number of tokens in the training and test set of each dataset. Since we don't test on Brown, this entry is left empty.}
\label{fig:datasetsize}
\end{figure}

\section{Additional Results}

\begin{figure}[h!]
    \centering
    \begin{tabular}{rrr}
    & \multicolumn{2}{c}{\textbf{Test Perplexity}} \\ 
    \cmidrule(lr){2-3}
    $(\bm{\sigma^2},\bm{\epsilon})$ & \textbf{Small} & \textbf{Large}  \\ \hline
    $(0.1,107.30)$ & $1480.84$ & \changed{1627.56} \\ 
    $(1.1,9.75)$ & $1473.49$ & NA \\ \hline
\end{tabular}
    \caption{We provide the trade off between $\epsilon$ and test perplexity for the small and large models from figure \ref{fig:exps_small_large}.  We hold $\delta$ to $1e-5$ and set the gradient clipping to $1.0$. We include the \changed{lowest} test perplexity for each model.  Recall the large model with $\sigma=1.1$ never converged to finite perplexity and is denoted NA.}
    \label{fig:epsilon_vs_perplexity}
\end{figure}

\begin{table*}
\begin{tabular}{ll|ll|ll}
\textbf{Training / Testing Set}   & $\bm{\sigma^2}$ & \textbf{PP (dev)} & \textbf{PP (test)} & \textbf{PP (dev, large)} & \textbf{PP (test, large)} \\ \hline
Brown / Reddit\_10k      & 0                      & 1561.20  & 1584.54 & 1652.65 & 1677.42 \\
Reddit\_10k / Reddit\_10k & 0                      & 3805.83  & 3787.68 & 1254.48 & 1259.23  \\

fine-tuned / Reddit\_10k & 0.0 & 1035.45 & 1037.81 & 1016.65 & 1019.31 \\ 
fine-tuned / Reddit\_10k      & 0.1                    & 1457.94  & 1480.84 & 1604.42 & 1627.56 \\
fine-tuned / Reddit\_10k      & 1.1                    & 1450.01  & 1473.48 & inf & inf
\end{tabular}
\centering

\caption{Final test-set perplexities for each of our models. Fine-tuned refers to the model being trained on Brown, then fine-tuned on the Reddit 10K training set. PP marked as ``large" are from the second, larger neural network we trained.  Note that $\sigma^2=0.0$ refers to a non-DP model while $\sigma^2 > 0.0$ is a DP model, where the privacy guarantee increases with $\sigma^2$.}
\label{tab:finaltessetperplexities}
\end{table*}

\begin{table*}
\begin{tabularx}{\textwidth}{llX}
 \textbf{Model}  & \textbf{Prompt} & \textbf{Sentence} \\ \toprule
 Reddit\_10k / Reddit\_10k $\sigma^2=0.0$& ``Bob lives close to the" & ``know extent better though about really said breaking will" \\
 fine-tuned / Reddit\_10k $\sigma^2=0.0$ & ``Bob lives close to the" & ``few alone saw good up done could branch clever been" \\
fine-tuned / Reddit\_10k $\sigma^2=0.1$ & ``Bob lives close to the" & ``city plans increase whose even reached years relieved construed what." \\ 
fine-tuned / Reddit\_10k $\sigma^2=1.1$ & ``Bob lives close to the" & ``along supply am certain like alone before decent exceeding other" \\
Large Reddit\_10k / Reddit\_10k $\sigma^2=0.0$ & ``Bob lives close to the" & ``above twice wanted therefore while unless however defective."\\
Large fine-tuned / Reddit\_10k $\sigma^2=0.0$ & ``Bob lives close to the" & ``once obviously give found now re like exact dislike out."\\
Large fine-tuned  / Reddit\_10k $\sigma^2=0.1$ & ``Bob lives close to the" & ``leaders kid forward governor thought neck let rides orchestral should" \\
fine-tuned GPT-2 / Reddit\_10k & ``Bob lives close to the" & ``station and we only have two miles of travel left to go" \\ 

\bottomrule
\end{tabularx}
\centering
\caption{A selection of sentences generated from the prompt ``Bob lives close'' using models finetuned on the Reddit $10k$ data set.  Except for the GPT-2 model, there's not a strong difference between the coherency of sentences generated. }
\label{tab:generatedsentences}
\end{table*}

\end{document}